\documentclass{article}
\usepackage{spconf,amsmath,graphicx,hyperref}
\usepackage{url}
\usepackage{booktabs,multirow, amsfonts}
\usepackage{xcolor}
\usepackage[most]{tcolorbox}
\usepackage[normalem]{ulem}

\title{Inference-time Target Speaker Unlearning in LLM-based \\Automatic Speech Recognition}
\name{Bo Su, Yueru Yan, Thai Le}

\address{
Indiana University, Bloomington, USA
}

\begin{document}
%
\maketitle
\begin{abstract} %
We introduce target-speaker unlearning ASR (TSU-ASR) task in a fully end-to-end framework for multi-speaker ASR and diarization. Given a multi-speaker utterance and a set of opt-out speakers who do not wish to have their speech transcribed, the task requires an ASR system to transcribe all speakers except the opt-out ones, while still indicating when those speakers are active. As a first step towards tackling this task, we introduce a novel, light-weight Enrollment-Conditioned Gating (ECG) module attachable to a frozen dual-stream speech LLM that enables ASR for new opt-out speakers dynamically during inference, even those who were not seen during initial ECG training phase. Our experiments on both AMI (English) and AliMeeting (Mandarin) datasets show that speech transcription accuracy for corresponding opt-out words or characters falls from 72.3\% to 48.2\% and from 73.6\% to 27.3\%, respectively, while retained speakers' transcription error rates maintain more or less the same. Our approach provides a practical solution for modern video conferencing platforms, allowing speakers to dynamically opt-out from automated AI transcriptions without forcefully leaving the meeting sessions, enabling a privacy-preserving interface for potentially millions of online meetings daily.
\end{abstract}

\section{Introduction}
\label{sec:intro}
AI transcription for online conferencing is often provided to automatically transcribe spoken conversations into text, often via an automated speech recognition (ASR) model. However, some participants may opt out for such an automated process~\cite{selectivehearing} or do not consent to have their speech transcribed. Unfortunately, there is currently no mechanisms from popular conference platforms that allows the AI transcription to ``ignore'' a few selected speakers. Post-recording solutions such as speaker-specific utterance or audio segments filtering do not solve the problem because not only it is after-the-fact, but also it can lead to unintended removal of other speaker's utterances when there are overlap in speech. Pre-deployment solutions such as retraining the models is both computationally expensive and impractical because such training process is costly~\cite{slamllm,tagspeech,speakerlm,jedis}, especially for recent advanced transcription system that is built upon complex LLM architectures, and opt-out requests can come not only post-deployment but also dynamically during inference. 

We formulate this task as \textit{target-speaker unlearning for ASR} (\textbf{TSU-ASR}). Unlike target-speaker ASR, which aims to transcribe only a selected speaker~\cite{tsasr_virtualspk,ssl_multitalker}, TSU-ASR aims to transcribe all speakers except those in an opt-out list. Intuitively, given a meeting recording and short voice samples of the speakers to exclude, the system should omit their words while preserving other speakers' transcripts. For the sake of integrity of online conferencing, TSU-ASR would also report who spoke when, known as speaker diarization, so readers can distinguish omitted speech from silence. For instance, the transcript could mark that \textit{Speaker A} spoke from 10 to 15 seconds without displaying their words. 

To tackle this problem, one branch of research that is applicable is machine unlearning. While existing machine unlearning approaches for AI speech models are limited, most of which often address a very different problem: removing the influence of training examples~\cite{mmunlearnsurvey,unlearnasr25,speechunlearning25}. In our case, we aim to exclude a selected speaker's utterances from the transcription, even if the ASR model has never been trained on that speaker's voice. Methods that suppress content from a model's inference stream such as ~\cite{eco24,guardit,trus} are more applicable, but doing so for meeting transcription must also handle changing speakers and overlapping speech, making our proposed problem both practical and non-trivial to tackle.

Therefore, in this work, we propose a novel module, called \textit{Enrollment-Conditioned Gating (\textbf{ECG})}, that can plug-and-play to an existing LLM-based speech model~\cite{tagspeech}, which has became increasingly more popular in the ASR literature. Existing LLM-based ASR models often process speech content and speaker identity through two separate paths. Intuitively, ECG then the voice samples to estimate where the selected speakers are speaking and scramble the information passed through the content path at those times, leaving the speaker path unchanged. After plugged to the model, ECG module is briefly trained for adaptation with the the original, backbone model fixed. This makes the selection of a new speaker to exclude to require no further training.

Our novel plug-and-play ECG module for TSU-ASR achieves a significant reduction in correct transcriptions of opt-out speakers when tested on datasets of meeting recordings in both English and Mandarin while observing \textit{no} noticeable degradation on transcription quality of the remaining speakers.

\vspace{-10pt}
\section{Problem Formulation}

Let $x$ be a recording containing a set of speakers $\mathcal{S}(x)$. Let $\mathcal{F}$ be the forget set or set of opt-out speakers, to be represented by one short enrollment audio $e_s$ per opt-out speaker. Let $\mathcal{R}(x)=\mathcal{S}(x)\setminus\mathcal{F}$ be the remaining, retained speakers. A TSU-ASR system computes $\hat{y}{=}f\big(x,\{e_s\}_{s\in\mathcal{F}}\big)$ in a single pass, with $\mathcal{F}$ chosen at inference, under two conditions: \textbf{(1) Opt-out speakers unlearned:} No utterances of $\mathcal{F}$ appear in $\hat{y}$
under any speaker label, while their diarization is retained, or the
resulting transcript must still report that someone spoke, and when; 

\textbf{(2) Retained speakers preserved:} On $\mathcal{R}(x)$, $\hat{y}$ matches model's outputs without TSU, including where target and retained speech might overlap.

The two conditions are independently satisfiable where the opt-out speaker talks alone, and compete where speech overlaps, since a mechanism acting on the time axis cannot suppress one voice in a frame without touching the other. This is also an intriguing problem to solve for an end-to-end multi-speaker LLM-based ASR framework. 

\section{Proposed Method}\label{sec:method}

We propose \textit{Enrollment-Conditioned Gating (\textbf{ECG})}, a trainable, plug-and-play module for an end-to-end multi-speaker LLM-based ASR framework. ECG uses short voice samples from opt-out speakers to scramble or mask the information used to transcribe their words, while leaving the other speakers' utterance stream unchanged.

\begin{figure}[t]
\centering
\includegraphics[width=0.95\columnwidth]{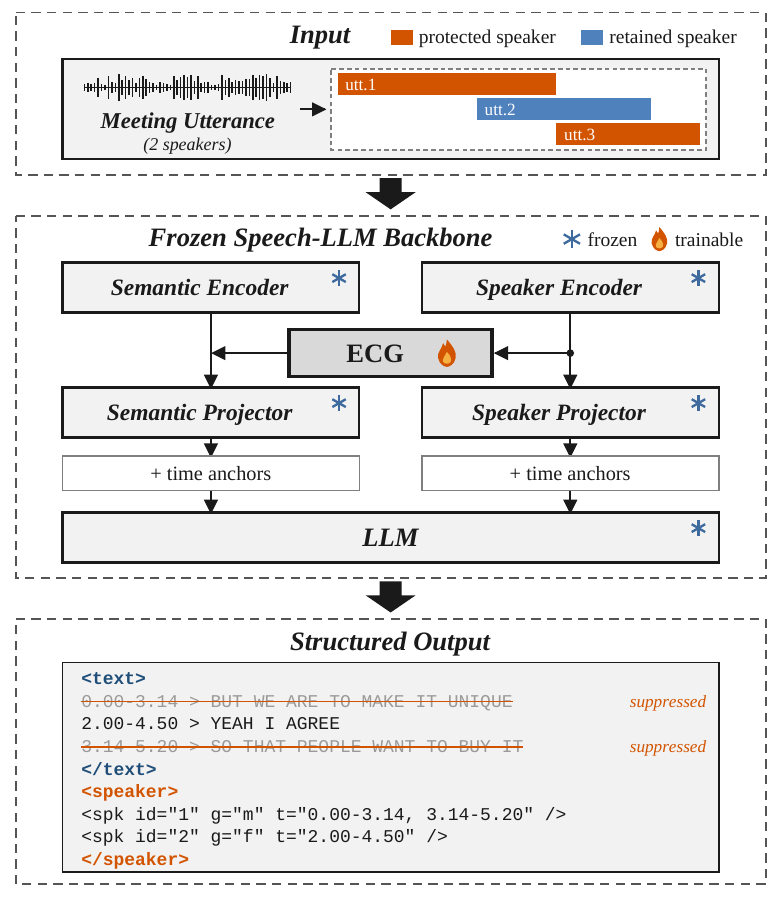}
\caption{Overview of TSU-ASR and the proposed ECG module that reduces information in the content stream during inference when target speakers are likely to be speaking. The speaker stream remains unchanged and provides information about who spoke when.}
\label{fig:method}
\end{figure}

\noindent \textbf{Base ASR Model.} Our work targets ASR architecture TagSpeech~\cite{tagspeech}, which uses two Zipformer encoders: a semantic encoder for speech content and a voice encoder for speaker information. Each encoder has a projector that converts its output into input features for a Qwen2.5-7B LLM. The LLM produces text, speaker labels, and speaking times in the XML format shown in Fig.~\ref{fig:method}. We use the publicly available checkpoints with all parameters frozen.

\noindent \textbf{Enrollment-Conditioned Gating Module.}
ECG estimates whether each frame, a short time segment of audio, contains speech from a opt-out speaker. First, frame-level voice features and enrollment embeddings are computed from the same voice encoder before projection, so they share the same feature space. Each enrollment embedding is obtained by averaging and normalizing features extracted from a opt-out speaker’s short voice samples. Then, for each frame, a light-weight two-layer multilayer perceptron (MLP) computes a matching score from the frame voice steam feature $v_t$ at time step $t$, an enrollment embedding $e_k$ of target speaker $k$, their element-wise product, and their cosine similarity:
\begin{align}
g_{t,e_k}
&= \sigma\!\Big(
\mathrm{MLP}\big[
v_t;\, e_k;\, \cos(v_t,e_k)
\big]
\Big),  
\label{eq:gate}
\end{align}
where $\cos(\cdot)$ denotes cosine similarity function, and $\sigma(\cdot)$ denotes logistic function returning a score in [0,1]. When multiple opt-out speakers are provided, we use the largest matching score at each frame. The matching score is then used to suppress the respective semantic stream $s_t$ at time step $t$, or:
\begin{align}
    s_t \leftarrow s_t * \max \{g_{t,e_k} \;|\;e_k \in \mathcal{F} \},
\end{align}
\noindent assuming that $\mathcal{F}{\neq}\emptyset$. Thus, higher scores lead to stronger suppression of the content stream. At the same time, ECG keeps the speaker stream unchanged, retaining the information used for speaker diarization. The gate uses continuous scores in [0,1] rather than binary decisions for easy of training through back-propagation. The set of opt-out speakers can be then changed by providing different voice samples, without further training.

\noindent \textbf{Training and Inference.}
We train only the ECG module while keeping the base model's parameters frozen. With probability 0.5, we select speakers present in the recording as targets and remove their words from the training transcript, while keeping the speaker labels and speaking times unchanged. Otherwise, no speakers are selected for exclusion and the original transcript is used. Each training example also includes a voice sample from a speaker absent from the recording as a negative example.

Our training loss $\mathcal{L}$ combines next-token cross-entropy on the desired output y against y- the reference with the forget speakers' text removed and frame-level binary cross-entropy on the gate logits g, using binary label m indicating speaker-matching segments as follows:
\begin{equation}
  \mathcal{L} = \alpha\mathrm{CE}(\hat{y},\, y^{-}) + \beta\mathrm{BCE}(g,\, m),
  \label{eq:loss}
\end{equation}
\noindent where $\alpha, \beta$ are coefficients to balance the two loss terms. The first term helps guide the generated output, while the second term helps optimize speaker matching ability.

\section{Experiments}

\subsection{Datasets and Experimental Setup}

We evaluate ECG on both AMI~\cite{ami} (English, single distant microphone) and AliMeeting~\cite{alimeeting} (Mandarin, first far-field channel). 
Following the base model's decoding settings, we keep each speech segments lasting 0.5--80\,s. This leaves 2{,}607 segments on AMI and 4{,}373 on AliMeeting. 

For each dataset, we first curate the test split by choosing roughly 10\% of all recording hours, resulting a set of 16 speakers for AMI and 60 speakers for AliMeeting. We use the remaining data as train split, making sure that there are no overlap in speakers between train and test. This allows us to test the generalizability of ECG beyond speakers seen during training phase. 

For each test split, we randomly sample 25\% of the speakers (4 out of 16 for AMI and 15 out of 60 for AliMeeting) as set of opt-out speakers, and the remaining 75\% serves as the retain set. We repeat this sampling 10 times and average the results. We collect five voice samples of 3--8\,s per opt-out speaker, using intervals where that speaker talks alone. During testing, we average the embeddings from all five samples to compute $e_k$ for each target speaker $k$. 

We report results for four segment types: \textit{Retain-only}, containing only retained speakers, \textit{Forget-only}, containing only opt-out speakers, and two mixed types containing both groups. \textit{Mixed-overlap} includes segments where protected and retained speech overlap for more than 0.05\,s. The remaining mixed segments are \textit{Mixed-nonoverlap}. Table~\ref{tab:testset} reports their sizes and durations for test split. Mixed-nonoverlap segments are uncommon because segments are split at pauses. All results are reported with $\alpha\leftarrow1.0$ and $\beta\leftarrow1.0$. 

\subsection{Evaluation Metrics}
\label{sec:metrics}

We evaluate opt-out speaker content leakage and retained speaker transcription accuracy separately. Utterances are lowercased and punctuation is removed. We score AMI in word level and AliMeeting in character level.

\noindent \textbf{Content Leakage Rate.} CLR measures how much of the opt-out speakers' reference utterance remains anywhere in the generated transcript:
\begin{align}
\mathrm{CLR}
=
\frac{\sum_x \mathrm{LCS}\big(r(x),h(x)\big)}
     {\sum_x |r(x)|},
\end{align}
where $r(x)$ is the reference text spoken by the opt-out speaker in segment $x$, and $h(x)$ is the full generated transcript, and $LCS(\cdot)$ computes the length of the longest common subsequence: it counts the largest number of matching words or characters in the same order, without requiring them to be consecutive. We report \textit{CLR-all} using all words or characters and \textit{CLR-rare} for evaluating on only rare words or characters or those with document frequency below 1\%. Restricting the measure to rare units reduces matches caused by common expressions shared across speakers. Both scores are reported as percentages, with lower values indicating effective suppressing. Because CLR examines the full transcript, opt-out content still counts if it is assigned to a retained speaker.

\begin{table}[tb!]
\centering
\caption{Test segments by type, averaged over ten protected-speaker sets. Dur.: total duration; Ovl.: overlap between protected and retained speakers (both in hours).}
\label{tab:testset}
\small
\setlength{\tabcolsep}{2pt}
\begin{tabular}{lrrrrrr}
\toprule
& \multicolumn{3}{c}{AMI-SDM}
& \multicolumn{3}{c}{AliMeeting} \\
\cmidrule(lr){2-4}
\cmidrule(lr){5-7}
 & \# & Dur.(h) & Ovl.(h) & \# & Dur.(h) & Ovl.(h) \\
\midrule
Retain-only      & 1,506 & 3.28 & --   & 2,592 & 5.24 & --   \\
Mixed-nonoverlap & 11    & 0.01 & 0.00 & 15    & 0.03 & 0.00 \\
Mixed-overlap    & 732   & 3.07 & 0.47 & 938   & 3.19 & 0.53 \\
Forget-only      & 359   & 0.56 & --   & 827   & 1.02 & --   \\
\midrule
Total            & 2,607 & 6.92 & 0.47 & 4,373 & 9.48 & 0.53 \\
\bottomrule
\end{tabular}
\end{table}

\begin{table*}[ht]
\centering
\small
\setlength{\tabcolsep}{5pt}
\caption{Each cell depicts score changes from \textit{without} to \textit{with our module attached}, averaged over ten random draws of the retained speakers. Lower is better throughout.}
\label{tab:main}
\begin{tabular}{lcccccccc}
\toprule
Dataset & \multicolumn{4}{c}{AMI-SDM} & \multicolumn{4}{c}{AliMeeting} \\
\cmidrule(lr){2-5}\cmidrule(lr){6-9}
& \multicolumn{2}{c}{Forget side} & \multicolumn{2}{c}{Retain side}
& \multicolumn{2}{c}{Forget side} & \multicolumn{2}{c}{Retain side} \\
\cmidrule(lr){2-3}\cmidrule(lr){4-5}\cmidrule(lr){6-7}\cmidrule(lr){8-9}
        & CLR-rare$\downarrow$ & CLR-all$\downarrow$ & cpWER$\downarrow$ & gWER$\downarrow$
        & CLR-rare$\downarrow$ & CLR-all$\downarrow$ & cpCER$\downarrow$ & gCER$\downarrow$ \\
\midrule
Retain-only      & ---               & ---               & 36.7\,$\to$\,37.0 & 32.7\,$\to$\,32.9
                 & ---               & ---               & 45.7\,$\to$\,45.6 & 43.6\,$\to$\,43.4 \\
Mixed-nonoverlap & 56.0\,$\to$\,32.5 & 78.2\,$\to$\,61.7 & 55.5\,$\to$\,59.9 & 54.9\,$\to$\,58.9
                 & 83.6\,$\to$\,28.4 & 90.6\,$\to$\,35.3 & 18.6\,$\to$\,45.7 & 18.6\,$\to$\,44.5 \\
Mixed-overlap    & 62.1\,$\to$\,41.5 & 67.4\,$\to$\,48.2 & 62.7\,$\to$\,65.6 & 56.7\,$\to$\,59.3
                 & 57.7\,$\to$\,28.7 & 64.7\,$\to$\,33.3 & 56.9\,$\to$\,65.5 & 53.9\,$\to$\,61.7 \\
Forget-only      & 80.4\,$\to$\,42.1 & 86.0\,$\to$\,48.1 & ---               & ---
                 & 80.2\,$\to$\,11.8 & 89.0\,$\to$\,16.9 & ---               & ---               \\
\midrule
All              & 67.0\,$\to$\,\textbf{41.6} & 72.3\,$\to$\,\textbf{48.2} & 48.9\,$\to$\,50.1 & 43.8\,$\to$\,44.8
                 & 65.4\,$\to$\,\textbf{22.7} & 73.6\,$\to$\,\textbf{27.3} & 49.1\,$\to$\,51.4 & 46.7\,$\to$\,48.7 \\
\bottomrule
\end{tabular}
\end{table*}

\noindent \textbf{Retained-Speaker Transcription Accuracy.}
We report standard word error rate metrics for meeting transcription~\cite{meeteval}, including \textit{cpWER}-R and \textit{gWER}-R on AMI, and the corresponding character error rates, cpCER-R and gCER-R, on AliMeeting. The suffix R indicates that these scores are calculated for retained speakers. We compare results with and without ECG to measure changes in transcription accuracy. 

For cpWER-R, we jointly match all reference speakers, both opt-out and retained, to predicted speaker labels using the assignment with the lowest total transcription error. We then score only the outputs matched to retained speakers. 
Outputs matched to opt-out speakers are excluded from this score, and unmatched outputs are ignored.

We also report retained-speaker errors alongside CLR: low transcription error alone does not establish that protected content has been removed, while an empty transcript can achieve zero CLR by omitting everyone's words.

\subsection{Results and Analysis}
\label{sec:results}

\begin{figure}[t]
\centering
\includegraphics[width=\columnwidth]{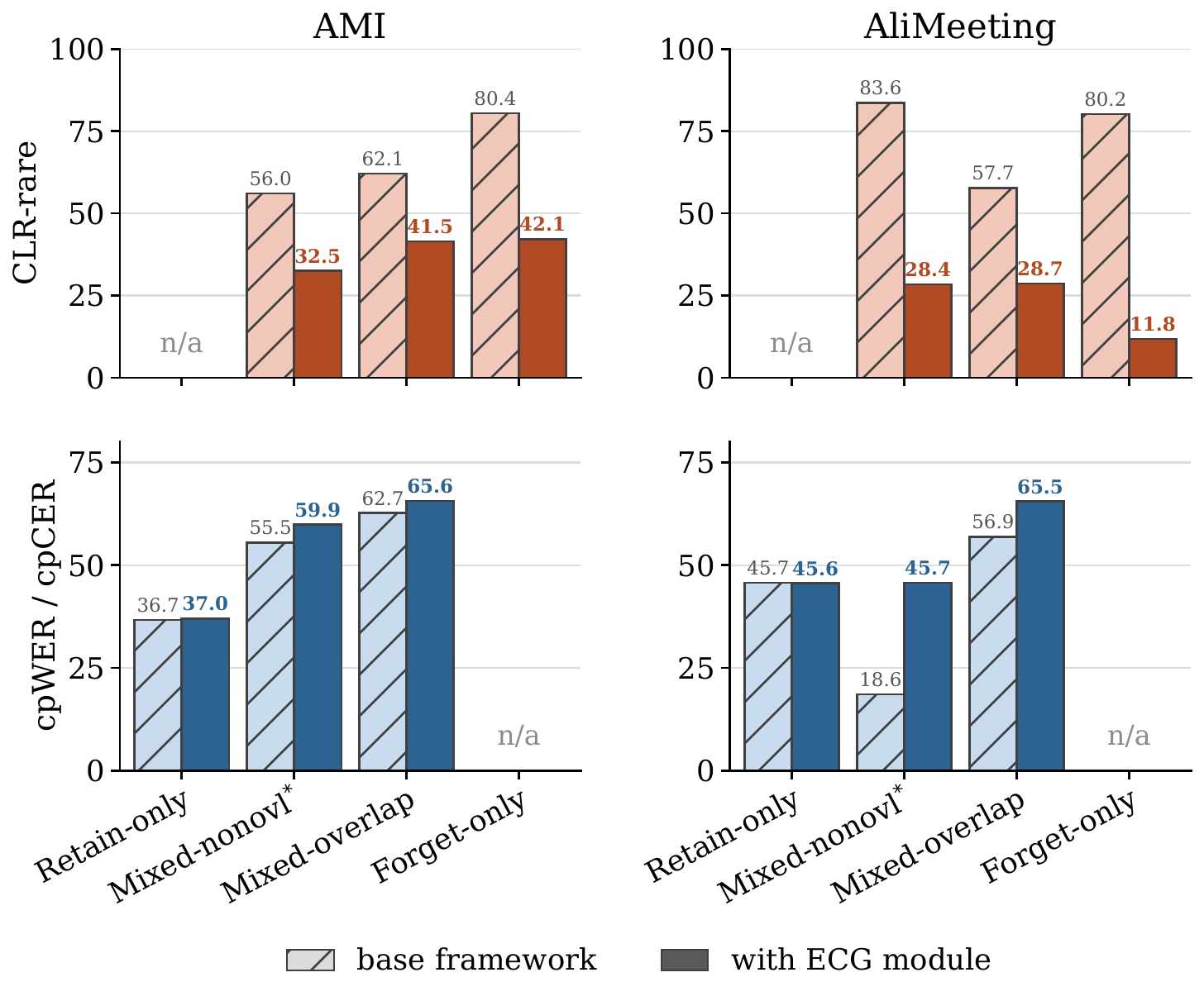}
\caption{CLR-rare (top) and retained-speaker transcription error (bottom), without and with ECG.}
\label{fig:strata}
\vspace{-8pt}
\end{figure}

\textbf{Overall Reuslts.} Table~\ref{tab:main} summarizes the results, part of which is visualized in Fig.~\ref{fig:strata} to show how the results vary across group types. Overall, ECG enables effective suppression of opt-out speakers during inference even if they were not seen during training, reducing CLR-rare by $25.4\%$ and $42.7\%$ in AMI and AliMeeting, while maintaining similar transcription performance in retained-speaker measured in cpWER and
cpCER (+$1.2\%$ and $2.3\%$). Noticeably, both CLR-rare and CLR-all consistently decrease across all groups containing opt-out speech, so the improvement remains when common words are included. 

\noindent\textbf{Group-level Results.} (Fig.~\ref{fig:strata}) show that group-level performance depends on which speakers are present in the segments with two main observations. Overall, the presence of retain speakers in the segments slightly hinders the suppression of opt-out speakers' transcriptions for AliMeeting but only marginally for AMI. Interestingly, opt-out content still leaks in Forget-only groups, even
though there is no retained speech to preserve. This shows that
overlap is not the only obstacle to removing protected words.
One possible explanation is that enough information still
reaches the frozen LLM to support transcription. 

Regarding retrained speakers, their error rates change little on Retain-only groups.
The loss in accuracy appears when protected and retained speech
share a group, where the model must suppress some content while
continuing to transcribe the rest. Errors increase in both
Mixed-overlap and Mixed-nonoverlap groups, although the small
sample size of the latter makes that comparison less significant. 

\noindent\textbf{Findings.} \textit{Finding \#1: Low leakage on Forget-only group alone is insufficient.}
An empty transcript achieves zero CLR, so a lower score does
not by itself establish successful selective removal.
For Forget-only groups, omitting all words is the intended
result, provided that the speaker information is preserved.
For mixed groups, the same output might also discard words
that should remain, making this a very challenging task. \textit{Finding \#2: the proposed ECG is a strong first step towards tackling TSU task,} demonstrating its ability to learn to match speaker patterns and translate it to a probabilistic gate that can control the residual semantic stream of the speech LLM architecture. Its simple implementation potentially allows adoption in other speech applications where LLM-based models are increasing prevalent and that we need to control the speech AI models dynamically with unseen conditional information such as opt-out speakers. 

\section{Conclusions}
We introduced a novel target-speaker unlearning for ASR task (TSU-ASR) that
aims to omit suppress speakers' utterances while preserving those of other
speakers. Our proposed Enrollment-Conditioned Gating (ECG) module suppresses the
semantic stream of an advanced multi-speaker LLM-based
ASR model while leaving its speaker stream unchanged. Our approach trains only the light-weight ECG, and require no further training for unseen opt-out speakers. Our experiments show significant effectiveness on two datasets AMI and AliMeeting in both English and Mandarin. 
\clearpage

\bibliographystyle{IEEEbib}
\bibliography{strings,refs}

\end{document}